# Analysis of Safe Ultrawideband Human-Robot Communication in Automated Collaborative Warehouse

Branimir Ivšić[1,2], Zvonimir Šipuš[2], Juraj Bartolić[2], Josip Babić[3]

[1] Ericsson Nikola Tesla d.d., Research and Development Centre, Radio Development Unit, Zagreb, Croatia, branimir.ivsic@ericsson.com *(from September 2019)*

[2] University of Zagreb, Faculty of electrical engineering and computing, Zagreb, Croatia, zvonimir.sipus@fer.hr, juraj.bartolic@fer.hr

[3] Končar–Electrical Engineering Institute Inc., Zagreb, Croatia, jbabic@koncar-institut.hr

*Abstract*—The paper presents the propagation analysis of ultrawideband Gaussian signal in an automated collaborative warehouse environment where human and robots communicate to ensure that mutual collisions do not occur. The warehouse racks are principally modeled as clusters of metallic (PEC) parallelepipeds, with dimensions chosen to approximate the realistic warehouse. The signal propagation is analyzed using a ray tracing software, with the goal to calculate the path loss profile for different representative scenarios and antenna polarizations. The influence of the rack surface roughness onto propagation is also analyzed. The guidelines for optimum antenna positions on humans and robots for safe communication are proposed according to the simulations results.

*Index Terms*—human-robot communication, UWB propagation in warehouse, rough surface reflection, safe communication.

## I. Introduction

Modern large warehouse systems and distribution centers are increasingly being operated by robots, which speeds up the process of warehouse management and enhances overall productivity. One of the most prospective automation concepts for large warehouses is based on the idea of having a fleet of wheeled mobile robots and modular portable racks [1, 2]. The robots are capable of traversing throughout warehouse both across the corridor (pathway) and under the racks, picking racks from the bottom and carrying them to the warehouse perimeter, where human can take the items and process them further. Given that modern warehouse dimensions can be of several hundreds of meters, that mobile robots can travel at speed of 1 m/s and carry racks weighing several hundred kilograms, the use of robots considerably reduces warehouse operation time and eases the job for humans [2].

To ensure smooth multi-robot operation in warehouse environment there is ongoing research in algorithms for optimization of the robot trajectories and ensuring continuous collision-free motion of the robots across the routes [3]. However, as humans still need to be capable to enter the warehouse (e.g. for service, maintenance or some intervention), safety issues arise since collisions between humans and robots need to be avoided. To guarantee safe warehouse operation for humans the most straightforward way is to separate the robots and humans completely from one another by e.g. solid or light fences as well as laser barriers around the working area of the robots [3]. This however results in the need to shut down the part of the system thereby increasing time of warehouse operation and limiting operation efficiency, especially in large warehouses. As a remedy to this issue, a new integrated collaborative warehouse paradigm [4] has arisen recently where humans and robots work closely together and move freely which gives rise to ensure safety by other means. Thus the first safety layer can be obtained by implementing into robot trajectories [3] human intention recognition algorithms [5, 6] which predict possible motion paths along the warehouse and reroute the robots.

The second safety layer in such collaborative environment is provided by equipping the robots with sensors to avoid collisions as well as using augmented reality technologies [4]. Finally, the third layer of safety can be achieved when human and robots additionally communicate, by which the human-robot proximity can be continuously evaluated, and robots can be stopped if the human is too close to them. This requires human to wear special equipment for communication such as the concept of SafetyVest [2] currently being developed. The SafetyVest is intended to be worn by all humans entering the co-working space and is equipped with radio ranging technology (antennas are placed both on the chest and the back), that facilitates implementing the safety architecture described above. The typical scenario of the automated collaborative warehouse is depicted in Fig. 1. Here one can note the robots carrying warehouse racks and stopping if they breach the safety distance to the human. The communication is intended to be realized using ultrawideband (UWB) signals, since they enable accurate localization, high data rates and are immune to fading which makes them suitable for use in multipath environment such as warehouse [7, 8].

To explore the SafetyVest communication limitations from the electromagnetic wave propagation viewpoint, in this paper we propose the model of the warehouse with modular rack clusters and perform ray-tracing analysis of propagation. That way we wish to calculate the path loss profile for several characteristic scenarios and gain an insight in the mechanisms of propagation and safe communication range in such environment. We also study the impact of different antenna polarizations as well as the impact of rough rack wall surfaces onto the signal propagation.

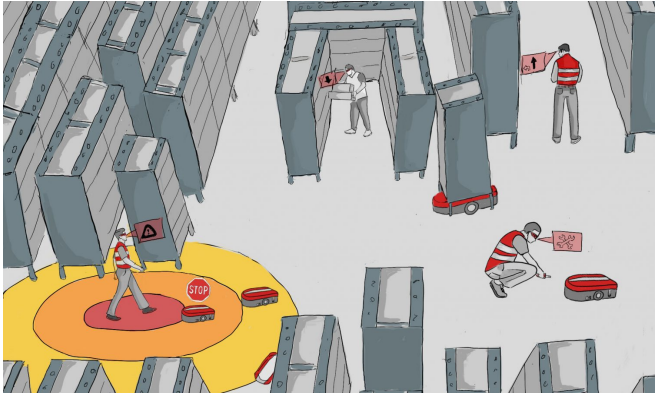

Fig. 1. A typical scene in automated collaborative warehouse [1]

## II. THE WAREHOUSE MODEL

For warehouse modelling and propagation calculations we use commercially available ray-tracing software Remcom Wireless Insite [9], with full 3D propagation model and shooting and bouncing ray method. The principal objects in the model and general problem formulation are illustrated in Fig. 2. We model the warehouse racks as clusters of parallelepipeds fully filled with metal (PEC), which represents the worst-case scenario as the propagation cannot occur directly through shelves. Thus the signal propagation behind racks is obtained through 30 cm high air gap between racks and floor (note that this gap in reality also serves as a possible movement path for robots as shown in Figs 1 and 2 [5, 6]). In simulation setup we allow for six reflections and first-order diffraction [9], by which most contributing rays are taken into account.

The transmitting antenna representing the antenna on Safety Vest is placed at the height of 1.5 m (which corresponds to an average human chest height), while the multiple receiving antennas (i.e. the antennas on robots) are placed as grid at 20 cm height. The horizontal dimensions of single rack are 1.3 m × 1.3 m while its height of is 2 m. The racks are grouped into clusters of 7×2 racks (Fig. 2), while between each two clusters the 1.5 m wide corridor is placed as shown in Fig. 3 (the corridor allows movement of both humans and robots). In addition, in the proposed model we set a small 5 cm gap between racks in clusters which also enhances communication (and is realistic). The warehouse floor is made of 30 cm thick concrete ($\varepsilon_r$=7; $\sigma$=0.015 S/m). Note that no walls are present in our warehouse model as we wish to minimize the number of possible reflection paths that are not attributed to the racks (by this we can also in principle generalize the problem to warehouse of any dimension).

In proposed model we consider four rack clusters (i.e. total number of 56 racks) giving the total observation area of 22 m × 8 m (Fig. 3). Such an area is comparable to the one in Safelog testing cell in Augsburg, Germany [2], however here we implement more stringent configuration in terms of rack number and material, to obtain the worst case that might occur in reality.

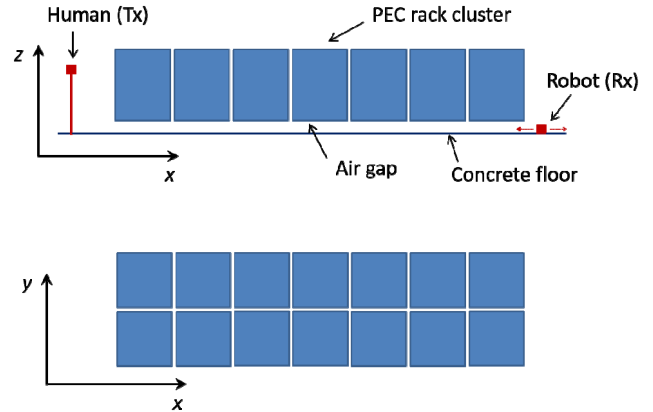

Fig. 2. The principle objects in the warehouse model to be analyzed (top: side view; bottom: top view of 7×2 rack cluster)

The communication between the human and robots is intended to be realized in UWB range by using Decawave DW 1000 transceiver specified in [10]. In particular, in our model we establish communication via an UWB Gaussian pulse centered at 3.994 GHz and 468 MHz wide, which corresponds to UWB Channel 2 [10]. The antennas to be used are planar circular wideband monopoles for which several commercial prototypes have been acquired and several prototypes additionally manufactured [11], all of which have been shown to possess adequate dipole-like radiation characteristics in the observed range. Based on the measurements in [11] for propagation model the antenna $E$-plane half-power beamwidth is taken as 60° with maximum gain of 3 dBi. Furthermore, the input power to the antenna is normalized to 0 dBm, which enables observing path loss profile. Since the receiver sensitivity is assumed to be -106 dBm, by observing relevant regulations on effective radiated power in UWB [10], we estimate the maximum allowed path loss for safe communication as up to around 80-90 dB.

## III. THE RECEIVED POWER PROFILES

Figure 4 shows the received power profile within the rack area of the warehouse when the transmitter (i.e. the human) is placed in the middle of the warehouse, while Fig. 5 shows the received power profile when the transmitter is placed in free space 2 meters before shelf cluster (the shelf clusters are shown in wireframe mode for convenience and to show signal propagation under the racks). Vertical polarization of all the antennas is assumed here. In both cases adequate signal coverage is obtained. For the case of Fig. 4 it can also be noted that the maximum signal is actually not directly under the transmitter, which is expected as this area is in the null of the supposed dipole radiation pattern, meaning that the direct signal component is absent there. In addition, slight asymmetry can be noted in the results as the simulation setup is not perfectly symmetric due to discretization of receiving grid.

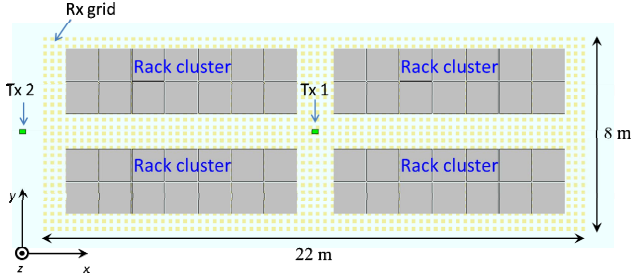

Fig. 3. The layout of the warehouse with four 7×2 rack clusters (top view)

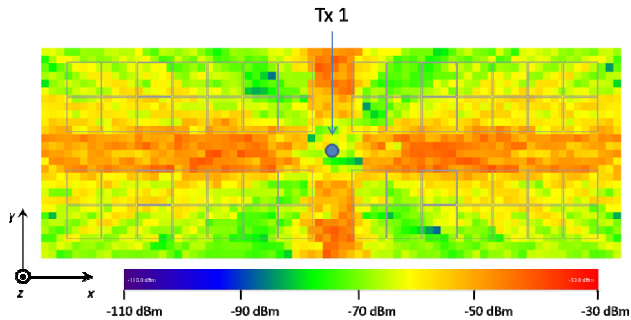

Fig. 4. Received power profile when the transmitter is placed in the middle of the warehouse

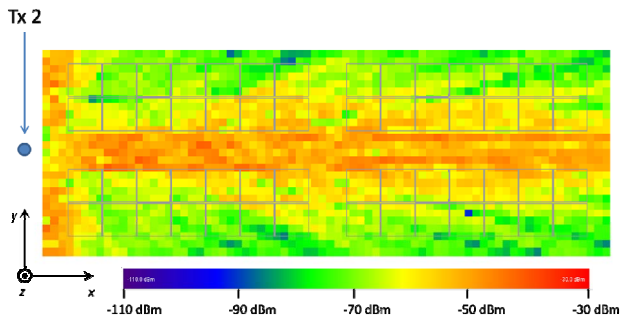

Fig. 5. Received power profile when the transmitter is placed before the leftmost cluster

## A. Other polarization setups

From the previous section it can be inferred that in the realistic warehouse even for the worst case scenario of PEC racks it is possible to establish safe communication to over 20 meters of distance regardless whether human is placed in the middle or at the end of the warehouse. Nevertheless, the next point of interest is to observe other possible polarization configurations in order to find out how much the antenna polarization affects propagation and the communication range. In Figs 6 – 8 the main results are shown for various relative orientations of transmitting and receiving antennas.

By comparing Figs 4 and 6 it can be noted that when the antennas on the robots are horizontally polarized more "blind spots" occur under racks. This is also seen by comparing Figs 7 and 8, where for horizontal polarization of robot antennas signal propagates rather poorly outside the corridor.

This difference in polarization response can be explained by noting that the horizontally polarized field under the racks is partially cancelled due to the image field arising from the boundary conditions at the rack bottom surface [12]. On the other hand, the signal from the vertically polarized antenna predominantly exhibits parallel ($H$-) incidence [12] with respect to the ground where the electric field has a significant vertical component which propagates under the racks better than its horizontal counterpart. The extent of this effect nevertheless needs further study and quantification in the future as the propagation mechanism in warehouse environment is rather complex to predict solely in deterministic terms [7].

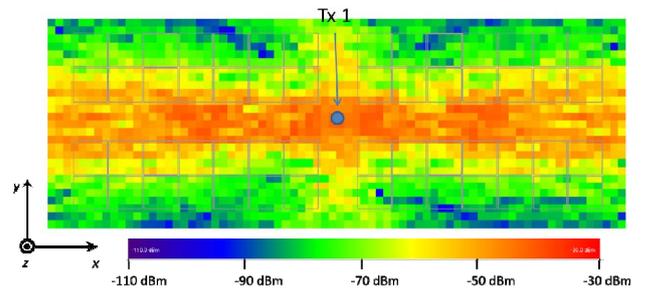

Fig. 6. Received power profile when both transmitter and receiver are horizontally polarized in transverse ($y$-) direction

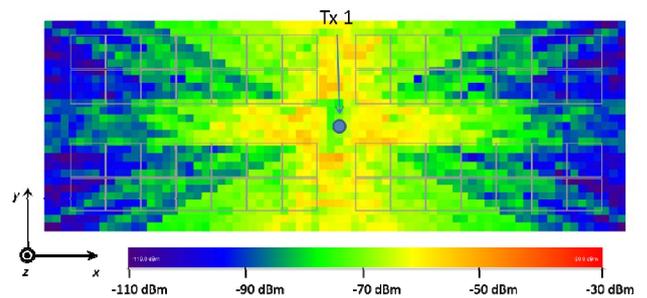

Fig. 7. Received power profile when the transmitter is horizontally polarized in $y$-direction while the receiver is vertically ($z$-) polarized

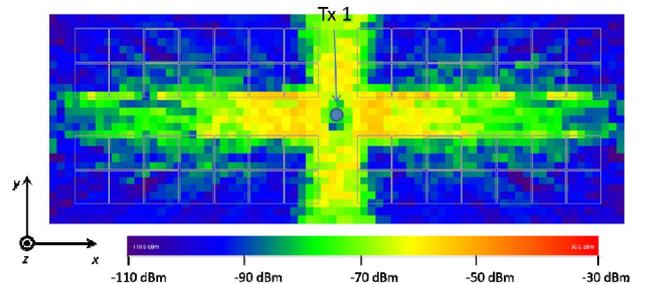

Fig. 8. Received power profile when the transmitter is vertically ($z$-) polarized, while the receiver is horizontally polarized (in $y$- direction)

## B. The "lying man" isssue

Another problem that arises in practice is that the human can change its body posture, which unlike robot movements cannot be predicted (he can e.g. kneel or even lie down on the floor if some warehouse task or service requires it, or if there is some accident). This of course changes the orientation (and hence polarization) of the transmitting antenna on the human body. To analyse this case, in subsequent simulation campaign we assume transmitter to be at height of 0.2 m which is the same height as robots (i.e. this is the case of lying human). The three possible transmitting antenna polarizations are analyzed in Figs 9 – 11 (the antennas on robots are assumed fixed with vertical polarization). It can be seen that very good signal coverage occurs when the transmitting antenna is vertically polarized (which is expected), while for other orientations the transmitted power is lower and can result in blind spots. This issue can be mitigated e.g. by mounting two antennas with different polarizations on the human body (i.e. using polarization diversity) and is to be considered in our future work.

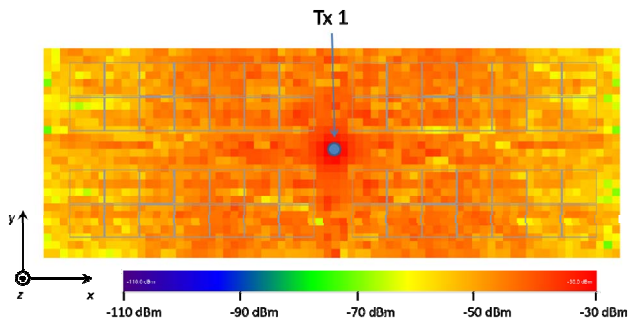

Fig. 9. Received power profile when both transmitter (placed at 0.2 m height) and receiver are vertically polarized

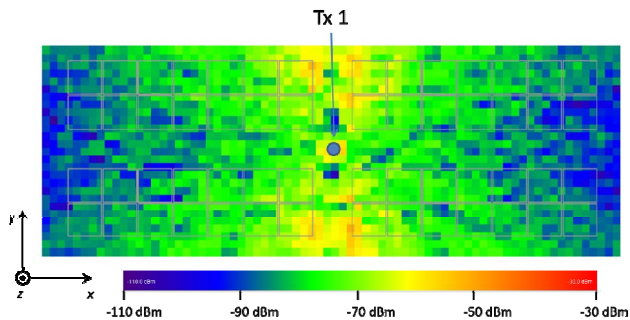

Fig. 10. Received power profile when the receiver is vertically polarized, while the transmitter (placed at 0.2 m height) is horizontally polarized in longitudinal (*x*-) direction

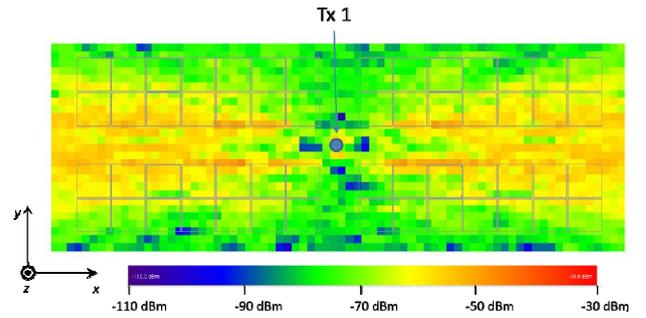

Fig. 11. Received power profile when the receiver is vertically polarized, while the transmitter (placed at 0.2 m height) is horizontally polarized in transverse (*y*-) direction

## IV. THE ROUGHNESS-CORRECTED RACK WALLS

In real warehouses the rack walls are not perfectly flat since the racks are typically unevenly filled with items, thereby affecting the waveguide effect across the corridor that has been observed in previous sections. We take this observation into account by allowing the rack wall surfaces to have random irregularities (deviations from flat surface) and by implementing into simulations the roughness corrected reflection coefficient *R* given as [13]:

$$R = R_0 e^{-8\left(\frac{\pi \cdot \Delta h \cdot \cos\theta_i}{\lambda_0}\right)^2}, \qquad (1)$$

where $R_0$ is the reflection coefficient for the flat surface, $\lambda_0$ is the wavelength, $\theta_i$ is the angle of incidence with respect to surface normal, while $\Delta h$ is the standard deviation of surface height around the mean height (the Gaussian distribution of height irregularities is assumed [13]).

For this scenario we perform another set of simulations for $\Delta h$= 5cm, by which we simulate rough rack walls, which is the amount of deviations that can be expected in reality. We note that since $\Delta h$ is comparable to wavelength the reflection coefficient (1) is actually an approximation [13], however for the purpose of our model it correctly enough predicts the ray behavior especially for larger angles of incidence.

For the case of racks filled with PEC ($R_0$=1) which we consider, this means (1) that reflection is significantly reduced for all angles of incidence less than around 80°, leaving grazing rays and diffracted rays as the major remaining mechanisms of propagation across the warehouse (i.e. the number of available rays is reduced compared to the case of flat rack walls). The results for the case when all the antennas are vertically polarized are given in Figs 15 and 16, for the two transmitter (human) positions. By comparing these results with the corresponding case of flat rack walls (Figs 4 and 5, respectively) we note that the received power in the corridors is reduced by around 20-25 dB which can nevertheless still be considered to ensure relatively safe communication.

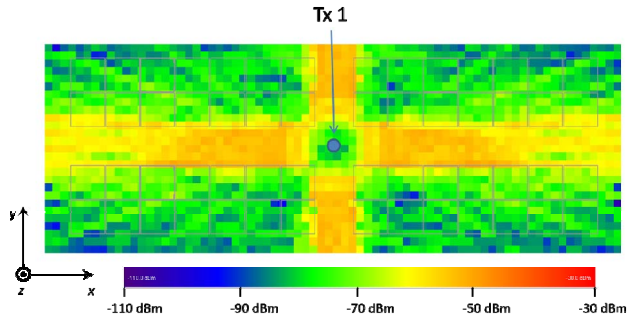

Fig. 12. Received power profile with roughness correction Δh = 5 cm (transmitter is placed in the middle of the warehouse)

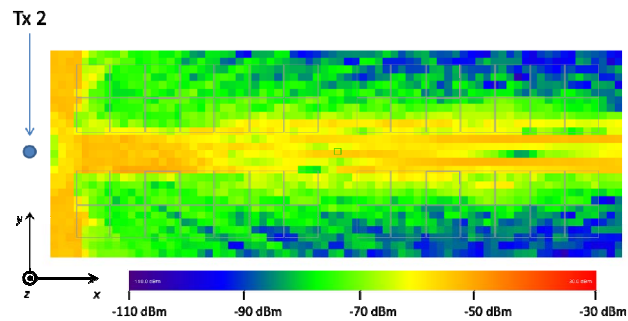

Fig. 13. Received power profile with roughness correction Δh = 5 cm (transmitter is placed before the leftmost cluster)

## V. Conclusion

In this paper we have proposed the collaborative warehouse model suitable for analysis of the propagation of electromagnetic waves using ray-tracing method. That way we have obtained fundamental limitations on communication range in warehouse environment, which can further serve as a guideline in logistic system design and estimation of the possible safe range of communication. In particular it was found that it is possible to obtain safe communication levels in most cases (by allowing the path losses up to 90 dB), while the optimum propagation is obtained when all the antennas are vertically polarized. The rough rack walls are shown to reduce the received power by around 20 dB.

The future work will be focused on the improvement of the theoretical propagation model and the evaluation of the extent of deterministic warehouse modelling. In addition, suitable laboratory measurement setup will be proposed, in order to verify the obtained results.

## Acknowledgment

This work has been supported from the European Union's Horizon 2020 research and innovation programme under grant agreement No. 688117 (SafeLog), and by Ericsson Nikola Tesla d.d. and University of Zagreb, Faculty of Electrical Engineering and Computing, under the project Emerging Wireless and Information Technologies for 5G Radio Access Networks (EWITA).